\documentclass{article}

\usepackage{arxiv}

\usepackage[utf8]{inputenc} 
\usepackage[T1]{fontenc}    
\usepackage{hyperref}       
\usepackage{url}            
\usepackage{booktabs}       
\usepackage{amsfonts}       
\usepackage{nicefrac}       
\usepackage{microtype}      
\usepackage{lipsum}
\usepackage{graphicx}
\usepackage{caption}

\usepackage{url}
\usepackage{dsfont}
\usepackage{xcolor}
\usepackage{mathtools}
\usepackage{amsmath}
\DeclareMathOperator*{\argmax}{argmax}
\usepackage{amssymb}

\usepackage[backend=bibtex, style=ieee]{biblatex}
\usepackage{latexsym}
\usepackage{framed,multirow}

\usepackage[bottom]{footmisc}

\newcommand*\samethanks[1][\value{footnote}]{\footnotemark[Joint senior author]}

\usepackage{latexsym}
\title{\Large Expectation Maximization Pseudo Labels}%

\author{
Moucheng Xu\thanks{contact: xumoucheng28@gmail.com. This manuscript has been accepted in Medical Image Analysis.} \\
UCL, London, UK \\ 
\And
Yukun Zhou \\
UCL, London, UK 
\And
Chen Jin \\
UCL, London, UK \\
\And
Marius de Groot \\
GSK, Stevenage, UK \\
\AND
Daniel C. Alexander \\
UCL, London, UK \\
\And
Neil P. Oxtoby\thanks{Joint senior author} \\
UCL, London, UK \\
\And
Yipeng Hu\footnotemark[2] \\
UCL, London, UK \\
\And 
Joseph Jacob\footnotemark[2] \\
UCL, London, UK 
}

\begin{document}
\maketitle

\begin{abstract}
In this paper, we study pseudo-labelling. Pseudo-labelling employs raw inferences on unlabelled data as pseudo-labels for self-training. We elucidate the empirical successes of pseudo-labelling by establishing a link between this technique and the Expectation Maximisation algorithm. Through this, we realise that the original pseudo-labelling serves as an empirical estimation of its more comprehensive underlying formulation. Following this insight, we present a full generalisation of pseudo-labels under Bayes' theorem, termed Bayesian Pseudo Labels. Subsequently, we introduce a variational approach to generate these Bayesian Pseudo Labels, involving the learning of a threshold to automatically select high-quality pseudo labels. In the remainder of the paper, we showcase the applications of pseudo-labelling and its generalised form, Bayesian Pseudo-Labelling, in the semi-supervised segmentation of medical images. Specifically, we focus on: 1) 3D binary segmentation of lung vessels from CT volumes; 2) 2D multi-class segmentation of brain tumours from MRI volumes; 3) 3D binary segmentation of whole brain tumours from MRI volumes; and 4) 3D binary segmentation of prostate from MRI volumes. We further demonstrate that pseudo-labels can enhance the robustness of the learned representations. The code is released in the following GitHub repository: \url{https://github.com/moucheng2017/EMSSL}.
\end{abstract}


\section{Introduction}

Recent years have witnessed the rise of deep learning based AI technologies in a wide range of applications for the betterment of humanity. The training of a successful deep learning model demands a large volume of annotated data. Regrettably, the money and time costs associated with the annotation acquisition is very expensive, causing a common issue namely label scarcity. The issue of label scarcity is especially challenging in one of the key applications of AI, healthcare, where the annotation process requires the expertise of highly skilled medical professionals, adding extra costs. In the era of AI-enabled healthcare, medical image segmentation is one of the core tasks, aiming at accurately labelling all pixels within volumetric medical images. It serves as a foundational step for other downstream tasks of healthcare, including computer-aided diagnosis, surgical navigation, and endpoint decision-making in drug discovery, among others. In this paper, we focus on the task of medical image segmentation as an exemplar application.

To tackle the inevitable issue of label scarcity in deep learning, semi-supervised learning has emerged as a solution. This approach utilises both labelled and unlabelled data to enhance model performance. Typically, unlabelled data are more abundant than their labelled counterparts, yet they are often overlooked in supervised learning paradigms. Semi-supervised learning is advantageous because it leverages existing unlabelled data, thereby sidestepping the need for additional investment in label acquisition. While other strategies, such as outsourcing data labelling combined with federated learning \cite{li_iclr_federated}, have been developed to address label scarcity, they still do not entirely eliminate the associated costs. In contrast, semi-supervised learning offers an attractive trade-off between cost and performance improvement. 

\subsection{Semi-supervised learning and Entropy regularisation}
\label{section:semi_supervised_learning}
Most semi-supervised learning methods focus on minimising the entropy of predictions for unlabelled data, aiming to make more ``firm" predictions. Entropy minimisation has a long-standing history in the field of representation learning. One of the earliest forms of this approach originated from the mutual information between input and output in unsupervised learning \cite{mutual_information_unsup}. The concept of entropy regularisation gained significant traction in the realm of semi-supervised image classification after it was proposed to minimise the entropy of unlabelled data as a strong form of regularisation \cite{EntropyMinimisation}. This technique aims to guide the model towards establishing a reliable decision boundary. Since its introduction, entropy minimisation has evolved from its original explicit form into various implicit manifestations.


\subsection{Consistency Regularisation}
Among the multifarious implicit forms of entropy minimisation, consistency regularisation stands as one of the most popular options, serving as the underpinning for a majority of cutting-edge methods in semi-supervised classification and segmentation \cite{fixmatch2020, remixmatch, bmvc_ssl_2018, bmvc_ssl_2020, Xu_MIDL}. Consistency regularisation utilises distance-based loss functions directly applied to the raw outputs or associated prediction probabilities. Consistency regularisation aims to engender predictive models that are resilient to perturbations at either the input or feature levels \cite{meanteacher, fixmatch2020, remixmatch, mixmatch, Xu_MIDL, cct_cvpr2020, bmvc_ssl_2020, cpl_cvpr_2021}.

For methods relying on input-level consistency, many are derived from a classic model called Mean-Teacher \cite{meanteacher}. In this model, the student's weight is an exponential moving average of the teacher model's weights. The teacher model processes a regular input, while the student model processes the same input with added Gaussian noise. In other words, the student model and the teacher model intake two different views of the same input. A mean square error is used for soft consistency regularisation between the outputs of the two models.

A more advanced teacher-student model, FixMatch, has achieved state-of-the-art performance in semi-supervised classification \cite{fixmatch2020}. FixMatch employs two forward passes: one with weakly augmented input (e.g., flipping) and another with strongly augmented input (e.g., shearing, random intensity). The output of the weakly augmented input is then used to generate a pseudo-label as the ground truth for training the output of the strongly augmented input.

Although FixMatch and its variants have excelled in image classification, it has been observed that they are not directly applicable to image segmentation tasks, as the cluster assumption does not hold at the pixel level \cite{bmvc_ssl_2020}. To adapt consistency regularisation for segmentation, the authors in \cite{cct_cvpr2020} found it feasible to apply perturbations at the feature level rather than the input level before implementing consistency regularisation. They directly apply augmentation techniques to the features of different decoders for semi-supervised image segmentation. Alternatively, perturbations can also be added through architectural modifications. For instance, one can train two identical models with different initialisations and apply consistency regularisation using pseudo-labels on both outputs \cite{cpl_cvpr_2021}. These methods, along with ours, are further tested and compared in a subsequent section \ref{section:results}.

\subsection{Pseudo Labelling}
Pseudo labelling is another form of entropy regularisation, requiring less computational resource. The concept of pseudo labelling was initially introduced in the context of semi-supervised multi-class image classification \cite{PseudoLabel}. In its prototypical form, pseudo labels are generated through the argmax operation applied to the output logits (essentially, the inferential outcomes) of the neural network for unlabelled data. Once generated, these pseudo labels are amalgamated with their corresponding unlabelled data and utilized to train the network in a manner akin to traditional supervised learning. One of the merits of this original approach lies in its computational efficiency; the generation of pseudo labels is performed ``on-the-fly," in real-time. It is common practice to initially ``warm-up" the network through purely supervised learning, followed by a gradual introduction—or ``ramp-up"—of the weight attributed to the pseudo labels in the loss function.

Pseudo labelling has garnered significant attention within both semi-supervised and self-supervised learning paradigms, chiefly owing to its computational frugality coupled with its robust performance metrics. Notably, some empirical studies have posited that semi-supervised learning strategies, leveraging pseudo labelling with vast volumes of internet-sourced unlabelled data, can outperform their fully supervised counterparts in tasks such as ImageNet classification \cite{meta_pseudo_label}. Recent research endeavors have sought to curtail the increasing complexity inherent in consistency regularisation techniques. These efforts have yielded competitive performance metrics, achieved solely through the judicious use of pseudo labels \cite{defense_pseudo_label}. Within the domain of image segmentation, novel methodologies underpinned by pseudo labelling have also demonstrated commendable results, especially when the pseudo labels are refined through self-attention mechanisms \cite{pseudoseg}.

However, pseudo labelling is not devoid of limitations; a notable issue is the phenomenon of confirmation bias \cite{confirmation_bias}. This occurs when erroneously generated pseudo labels are incorporated into the training process, thereby inducing a form of noisy training. The negative impact of these incorrect labels is not merely transient but tends to accumulate and amplify over the course of training. In the present manuscript, we propose a novel methodological framework aimed at mitigating this confirmation bias. Specifically, we introduce a stochastic training paradigm that is designed to learn the threshold of the pseudo labels, thereby generating high quality pseudo labels in an automatic manner.
\begin{figure*}[!h]
\centering
\includegraphics[width=\textwidth]{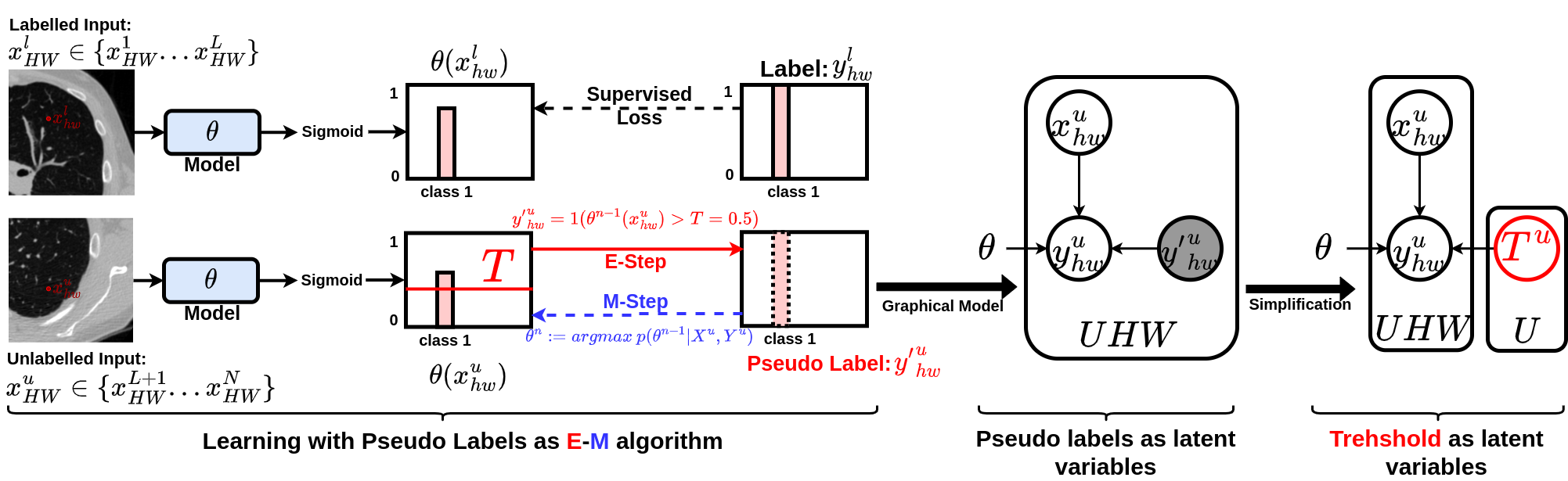}
\caption{Pseudo-labelling process for binary segmentation. Pseudo-label $y'_n$ is generated using unlabelled data $x_u$ and model with parameters from last iteration $\theta$. Therefore, pseudo-labelling can be seen as the E-step in Expecation-Maximization. The M-step updates $\theta$ using $y'_n$, $y$ and data $X$. In our 1st implementation, namely SegPL, the threshold $T$ is fixed for selecting the pseudo labels, which is the original pseudo labelling, as an empirical approximation of its true generalisation. In our 2nd implementation, namely SegPL-VI, the threshold $T$ is dynamic and learnt via variational inference, which is an learnt approximation of its true generalisation.}
\label{fig:main_method}
\end{figure*}

\subsection{Motivations and contributions}
It has come to our attention that the majority of extant literature on pseudo-labelling primarily adopts an empirical methodology, conspicuously omitting an investigation into the foundational mechanisms underlying its empirically observed efficacy. Motivated by this lacuna, we embarked upon a more in-depth examination of pseudo-labelling and ascertained its significant theoretical relationship with the classical Expectation-Maximisation (EM) algorithm in machine learning. Furthermore, our study is catalysed by contemporary research in the domain of semi-supervised image classification, which posits that achieving competitive performance metrics is feasible through judicious selection of high-quality pseudo-labels \cite{defense_pseudo_label}. In this manuscript, we offer a theoretical exegesis that elucidates the correlation between pseudo-labelling and the EM algorithm. Concurrently, we engage in empirical investigations to assess the applicability and robustness of pseudo-labelling in the context of semi-supervised medical image segmentation. We summarize our contributions in the following bullet points:
\begin{itemize}
    \item We interpret pseudo labelling as Expectation Maximization (EM) algorithm. As EM algorithm is gauranteed to converge to local minimum. We therefore partially explain the empirical success of pseudo labelling.
    \item We demonstrate the generalised form of pseudo labels. 
    \item We provide a learning method to learn the threshold for pseudo labelling in order to avoid confirmation bias and automatically pick up high quality pseudo labels.
    \item We investigate the use of pseudo labelling in semi-supervised medical image segmentation and its characteristics such as robustness.
\end{itemize}

Previously, a shorter version of this paper has been published at MICCAI 2022 \cite{xu_miccai_bpl}. This journal version includes a couple of extensions based on that MICCAI paper such as:
\begin{itemize}
    \item We fulfilled the details of the proposed probabilistic model of pseudo labelling.
    \item We expanded the results section and included one more data set of prostate segmentation from MRI volumes.
    \item We included the results on the whole data set of the BraTS 2018 which was only partially used before in the previous MICCAI version.
    \item We extended the related work section by including more recently proposed works.
    
\end{itemize}

\section{Related works}

The landscape of semi-supervised segmentation is, to a significant extent, influenced by the advancements in semi-supervised classification techniques, as delineated in Section \ref{section:semi_supervised_learning}. Among the various frameworks adopted, the mean-teacher based consistency regularisation paradigm is particularly prevalent in the field of semi-supervised medical image segmentation \cite{miccai_data_augmentation_2020, miccai_dual_teacher_2020, miccai_roi_consistency_2019, local_global_consistency_miccai_2020, pairwise_miccai2020, Ta_miccai2020, To_miccai2020, Unnikrishnan2020,Yang2020, Fotedar2020}. One early contribution to this vein of research was made by Yu et al., who enriched the mean-teacher model by incorporating uncertainty measures to generate a mask. This mask then modulates the application of consistency regularisation to only low-uncertainty regions \cite{miccai_uamt}.

Beyond perturbations at the data level, feature-level perturbations for consistency regularisation have also garnered considerable attention. For example, Luo et al. employed distinct initializations for different decoders to induce feature perturbations \cite{midl_luo_2022}. Xu introduced the idea by applying consistency regularisation on features after different morphological perturbations \cite{Xu_MIDL}. Another intriguing work employed a multi-decoder architecture, utilizing three decoders with divergent up-sampling layers, to enable mutual consistency regularisation across the decoder outputs \cite{WU2022102530}.

Recent studies have also explored the application of consistency regularisation to align signed distance maps of object boundaries, derived from different views of a common unlabelled input \cite{9740182}. The method is intended to enforce the models to be more of awareness of the object boundaries. Similar initiatives have leveraged uncertainty estimates, acquired via MC dropout, to weight the consistency regularisation \cite{ZHANG2023102476}. While these methods achieved good performances on some of the tasks, it is pertinent to note that the necessity for at least two forward passes substantially elevates the computational overhead. In contrast, our method does not bring extra computational burden and it requires minimalist changes of the original backbone segmentation model. 

Alternatively, more computationally affordable strategies have been pursued, notably leveraging pseudo labelling in the realm of medical image segmentation. Our proposed method also belongs to this paradigm. For instance, Bai et al. utilized conditional random fields to filter out false positives in pseudo labels \cite{ssl_cardiac}. Wang employed uncertainty measures to refine pseudo labels \cite{SSS_CT_Attention}. Wu et al. amalgamated pseudo labels with a dual-headed neural network architecture to instantiate a cross pseudo-supervision framework \cite{SSS_MCT}. Moreover, a recent study employed a variational auto-encoder as a student model to learn from pseudo labels generated by a deterministic teacher model \cite{wang_cvpr_2022_bayesian_ssl}. 


\section{Pseudo Labelling As Expectation-Maximization}
\label{sec:em_pl}

In this section, we reinterpret pseudo labelling in semi-supervised learning through the lens of the Expectation-Maximization (EM) algorithm. We specifically focus on binary segmentation, as it is commonly encountered in medical imaging tasks where the objective is to differentiate foreground from background. This framework can be easily extended to multi-class segmentation by employing a multi-channel Sigmoid function. Each channel is treated as a binary output and combined using the argmax operation for the final prediction.

\subsection{Problem formulation}
Given a set of $N$ total available training images as $X = \{x_n \in R^{HW} : n \in (1, 2, ..., L, L+1, ..., N)\}$, where $X_L = \{x_l \in R^{HW} : l \in (1, ..., L)\}$ are $L$ labelled images; $Y_L = \{y_l \in R^{HW} : l \in (1, ..., L)\}$ are $L$ labels for $X_L$; $X_U = \{x_u \in R^{HW} : u \in (L+1,..., N)\}$ is the rest of the $U$ or $(N-L)$ unlabelled images. We have a segmentation network with parameters as $\theta$ and our final goal is to predict the labels $p(Y|X, \theta)$ of the whole data $X$ with respect to $\theta$.

\subsection{Pseudo labels as latent variables} 
In order to find the optimal parameters of $\theta$, the common approach is maximum likelihood estimation for maximising the likelihood of $P(X | \theta)$ with respect to $\theta$, which contains two parts, namely supervised learning part and unsupervised learning part. The supervised learning part is to find the following joint data density with known full information of the labels:
\begin{equation}
    p(X_L, Y_L | \theta)
    \label{equ:sup_data_likelihood}
\end{equation}

The unsupervised learning part is to find the beneath likelihood with the same parameters without full information of the data:
\begin{equation}
    p(X_U| \theta)
    \label{equ:unsup_data_likelihood}
\end{equation}

Since labels are not observable for $X_U$, we can treat this as a missing data problem and introduce latent variables $Y'_U$. We therefore transform the above Eq.~\ref{equ:unsup_data_likelihood} to an estimation of the following marginal likelihood:
\begin{equation}
    p(X_U | \theta) = \int p(X_U, Y'_U | \theta) d Y'_U 
    \label{equ:unsup_marginal_likelihood}
\end{equation}

The latent variable in the above Eq.~\ref{equ:unsup_marginal_likelihood} can be implemented as the pseudo labels. Eq.~\ref{equ:unsup_marginal_likelihood} also shows that it is not an easy task to train a model in semi-supervised fashion, because it is difficult to simultaneously estimate the optimal values of $\theta$ and $Y'_U$. To address this difficult learning problem, we can decompose this problem by iteratively estimating the the latent variables $Y'_U$ and the model $\theta$. We now notice that this can be solved by a typical Expecation-Maximization (EM)\cite{bishop_ml} algorithm. By plugging the Jensen's inequality, one can iteratively refine the Evidence Lower Bound of the log likelihood of the data in Eq.\ref{equ:unsup_marginal_likelihood} (see details in later section \ref{sec:convergence}). 

\subsection{E-M Pseudo Labelling}
We now display each component of the pseudo labelling in the sense of EM algorithm in the following paragraphs.

\textbf{E-step} At the $n^{th}$ iteration, the E-step estimates the values of the latent variable with the model ($\theta^{n-1}$) from the last iteration ($n-1$). According to the cluster assumption that similar data points are supposed to have similar labels \cite{ssl_book}, the E-step runs the inference on unlabelled data and generate pseudo-labels according to its maximum predicted probability. In practice, in binary segmentation, the pseudo-labels for the foreground class 1 are picked using a fixed threshold value ($T$) between 0 and 1. Normally, this threshold is set up as 0.5. This binarization is actually equivalent to the plug-in principle \cite{EntropyMinimisation}, which is a common approach for estimating the posterior probability using an empirical estimation in statistics. Therefore, the pseudo-labelling itself is the E-step:

\begin{equation}
    y_u^{hw'} = \mathds{1} (\theta^{n-1}(x_u^{hw}) > T = 0.5)
    \label{equ:pseudo_label}
\end{equation}

The above Eq.~\ref{equ:pseudo_label} is pseudo-labelling at pixel-wise. Where $h$ and $w$ are the index for the height and the index for the width of the pixel location respectively, for each unlabelled image $x_u$. $y_u^{hw'}$ is the pixel-wise pseudo label. More details of the connection between E-step and pseudo labelling is in the later section sec.\ref{sec:convergence} on the convergence of pseudo labelling.

\textbf{M-step} At the M-step of iteration $n$, we will update the model parameters $\theta^{n-1}$ using the estimated latent variables (pseudo-labels $Y'_U$) from the E-step. The images $X$ are ignored for simplicity in the following expression:

\begin{equation}
    \theta^n := \argmax_\theta p(\theta^{n} | \theta^{n-1}, Y'_n)
    \label{equ:m_step_final}
\end{equation}

The above Eq.\ref{equ:m_step_final} is normally solved by setting the partial derivatives of the sum of the $p(Y'_n)$ with respect to $\theta$ as zero, which can be calculated with modern automatic differentiation based deep learning toolbox such as Pytorch \cite{pytorch}. In practice, we optimise $\theta$ in Eq.\ref{equ:m_step_final} via stochastic gradient descent. To use the stochastic gradient descent, we need to define an objective function and we use the common Dice loss ($f_{dice}(.)$) \cite{dice_loss} as this is a segmentation task:

\begin{equation}
    f_{dice}(a, b) = \frac{2 * a * b + \epsilon}{a + b + \epsilon} 
    \label{equ:dice}
\end{equation}

Where $a$ is the prediction, $b$ is the prediction and $\epsilon$ is to prevent the division of zero.

\textbf{Loss function of SegPL} We weight the Eq.\ref{equ:m_step_final} with a hyper-parameter $\alpha$. For the whole data set including both unlabelled and labelled data, we can extend the Eq.\ref{equ:m_step_final} and Eq.\ref{equ:pseudo_label} to a combination between the supervised learning part $L_L$ and the unsupervised learning part $L_U$:

\begin{equation}
\begin{split}
    & \mathcal{L}_{SegPL} = \underbrace{\alpha \frac{1}{N-L} \sum_{u=L+1}^{N} f_{dice}(\theta^{n-1}(x_u), \mathds{1} (\theta^{n-1}(x_u) > T=0.5))}_{\mathcal{L}_U} \\
    & +\underbrace{\frac{1}{L} \sum_{l=1}^{L} f_{dice}(\theta^{n-1}(x_l), y_l)}_{\mathcal{L}_L}
    \label{equ:pseudo_label_loss}
\end{split}
\end{equation}

The above loss function \ref{equ:pseudo_label_loss} is the key component of our first proposed semi-supervised segmentation method, omitting pixels' locations for simplicity, which is referred as SegPL (Segmentation with Pseudo Labels) in the paper. $\mathcal{L}_{L}$ works to prevent the networks falling into trivial solutions, trivial solutions happen when networks constantly predict one single class for all of the pixels. 

\subsection{On the convergence of Pseudo Labelling from the perspective of EM}
\label{sec:convergence}
In this section, we explain how semi-supervised learning with pseudo-labelling will always converges from the perspective of EM. We first define an objective function, the value of which we aim to increase. In our case, it would be the log data likelihood $log p(X_U)$. We also need to introduce a surrogate function $q(Y'_U)$ which is any arbitrary distribution over the latent varaible $Y'_U$. We follow \cite{bishop_ml} and display the lower bound of the log data likelihood in the form of the Free Energy:
\begin{equation}
\begin{split}
& log p(X_U) \coloneqq log \int p(X_U, Y'_U | \theta) d Y'_U \\
& \geq \int q(Y'_U) log \frac{p(X_U, Y'_U | \theta)}{q(Y'_U)} dY'_U \coloneqq \mathcal{F} (q (Y'_U), \theta)
\end{split}
\label{equ:free_energy_data}
\end{equation}

The functional free energy can be transformed back to the log data likelihood \cite{bishop_ml}:
\begin{equation}
    \begin{split}
    & \mathcal{F} (q (Y'_U), \theta)  = log p(X_U) - KL[ q(Y'_U)|| p(Y'_U| X, \theta)]
    \end{split}
    \label{equ:data_kl}
\end{equation}

In the E-step of the iteration $n$, the free energy is:
\begin{equation}
    \begin{split}
    & \mathcal{F} (q (Y'_U), \theta^{n-1})  = log p(X_U) - KL[ q(Y'_U)|| p(Y'_U| X, \theta^{n-1})]
    \end{split}
    \label{equ:e_step_free_energy}
\end{equation}

As KL can never be negative, the above Eq.\ref{equ:e_step_free_energy} has an upper bound. In order to reach that upper bound of the free energy at the $n^{th}$ iteration, we need to minimise $KL[ q(Y'_U)|| p(Y'_U| X, \theta^{n-1})]$. The KL distance has its minimum value at zero only if $q(Y'_U)$ is equal to $p(Y'_U | X, \theta^{n-1})$. Therefore, we can simply replace the arbitrary function of latent variable $q(Y'_U)$ as the current estimated posterior of the latent variable:
\begin{equation}
    q(Y'_U) = p (Y'_U | X_U, \theta^{n-1})
    \label{equ:surrogate_function}
\end{equation}

The above Eq.\ref{equ:surrogate_function} can be implemented as pseudo-labelling in Eq.\ref{equ:pseudo_label}. In other words, pseudo-labelling essentially maximises the free energy of the log data likelihood in the E-step.

Intuitively, the subsequent M-step is applying supervised learning to optimise the model parameters with pseudo labels which are produced from the precursory E-step. As supervised learning can be seen as maximium likelihood estimation, thereby, M-step increases the log data likelihood. In more details, we know that the log data likelihood in the M-step after updating the model $\theta$ is:
\begin{equation}
log p(X_U) = \mathcal{F}(q(Y'_U), \theta^{n-1}) + KL[ p (Y'_U | X_U, \theta^{n-1})|| p(Y'_U| X_U, \theta^n)]
\label{equ:log_data_likelihood_m_step}
\end{equation} 
The KL term in the above Eq.\ref{equ:log_data_likelihood_m_step} becomes positive as the posterior of the latent variable $Y'_U$ is different from its previous value. Together, it is easy to tell that the M-step increases the data log likelihood by at least the increased amount of the lower bound.

Up to this point, it is clear to see that, pseudo labelling (E-step) combined with supervised optimisation of model parameters (M-step) can never decrease the log likelihood of the data, leading to guaranteed convergence towards local optima. Similar conclusion was reported in the the original EM paper
\cite{depmster_em}. 


\section{Generalisation of Pseudo Labels via Variational Inference for Segmentation}
In the last section \ref{sec:em_pl}, we use an empirical estimation of the posterior of the latent variables (pseudo labels) by setting the $T$ as 0.5. The fixed empirical estimation of $T$ could be sub-optimal especially in the early stage of training when the networks do not have good representations and the predictions are not very confident \cite{defense_pseudo_label}. Potentially, noisy training with some ``bad" pseudo labels could accumulate some errors into the learnt representations. To address this potential issue, we provide an alternative approach to learn to approximate the true posterior of the pseudo labels. This alternative approach can be seen as a generalisation of the empirical estimation approach in SegPL in section \ref{sec:em_pl}. 


\subsection{Confidence threshold as latent variable} In the last section \ref{sec:em_pl}, we directly treat pseudo labels as latent variables. However, in segmentation task, the pseudo labels are pixel-wise, making the generative task a difficult one. To address this, we now introduce a simplification of the graphical model of the pseudo-labelling in \ref{sec:em_pl}. The key of this simplification is to treat the threshold value $T$ as the latent variable for instead: 
\begin{equation}
    p(X_U | \theta) = \int p(X_U, T | \theta) d T 
    \label{equ:unsup_marginal_likelihood_threshold}
\end{equation}
This new latent variable $T$ makes the computation of the posterior much easier. We know that $T$ is a value between 0 and 1, so that we have a clear prior knowledge of the range of this single value $T$. That any distribution describing values between 0 and 1 can be used as a prior distribution to approximate the real distribution of $T$. The true posterior of the latent variable $T$ is:  

\begin{equation}
\begin{split}
    p(T | X_U, \theta) = \frac{p(X_U |T, \theta)p(T)}{p(X_U | \theta)}
    \label{equ:e_step}
\end{split}
\end{equation}

The new E-step at iteration $n$ with threshold as the latent variable now becomes:

\begin{equation}
\begin{split}
    & p(T_n=i | X_U, \theta^{n-1}) = \\
    & \frac{\prod_{u=L+1}^{N} p(x_u |\theta^{n-1}, T_n=i)p(T_n=i)}{\sum_{j \in [0, 1]} \prod_{u=L+1}^{N} p(x_u |\theta^{n-1}, T_n=j)p(T_n=j)}
    \label{equ:e_step_full}
\end{split}
\end{equation}

From the above Eq.\ref{equ:e_step_full}, one can tell that the empirical estimation of the threshold $T$ is actually necessary although not optimal. Because there are infinite possible values between 0 and 1 in the denominator in Eq.~\ref{equ:e_step_full}, the posterior of the pseudo-labels is still intractable.

\subsection{Variational E-step} 
To address the aforementioned intractable issue in Eq.~\ref{equ:e_step_full}, we use variational inference for the approximation of $p(T)$. As mentioned before, the prior of $T$ can be an arbitrary distribution describing values between 0 and 1. For the implementation simplicity, we adapt an univariate Normal distribution for the prior distribution and we denote the prior distrinbution of T as a surrogate distribution $q(\beta)$. We use extra model parameters $\phi$ to parameterize the log variance and the mean of the approximated posterior distribution of $T$, conditioning on the image features, see the beneath Eq.~\ref{equ:pt_parametrization}. $\phi$ is implemented as a average pooling layer followed by a single 3 x 3 convolutional block including ReLU and normalisation layer, then two 1 x 1 convolutional layers for $\mu$ and $Log(\sigma^2)$ respectively. Alternatively, a simple fully connected layer can also be used as $\phi$, we found no performance differences among different choices of architectures for $\phi$.

\begin{equation}
    (\mu, Log(\sigma^2)) = \phi(\theta(X_U))
    \label{equ:t_model}
\end{equation}
\begin{equation}
    p(T|X_U, \theta, \phi) \approx  \mathcal{N}(\mu, \sigma)
    \label{equ:pt_parametrization}
\end{equation}

Differing from the fixed $T$ in E-step in Eq.~\ref{equ:pseudo_label}, the $T$ in variational E-step is dynamic, we denote the stochastic threshold as $\textcolor{red}{\textbf{T}}$ for clarity. We use the standard reparameterization trick \cite{VAE} to generate the threshold in each iteration:

\begin{equation}
\begin{split}
& \textbf{\textcolor{red}{T}} = \mu + rand*e^{0.5 * log(\sigma^2)} \\
& rand \sim \mathcal{N}(0, 1)
\end{split}
\label{equ:reparam}
\end{equation}

As demonstrated in previous Eq.\ref{equ:free_energy_data} that the log data likelihood term has an Evidence Lower Bound (ELBO) which contains a conditional probability of the data given latent variable and a KL distance between the posterior and the prior of the latent variable. We therefore write down variational unsupervised learning objective as: 
\begin{equation}
\begin{split}
    & Log (P(X_U) \geq  \\
    & \sum^{N}_{u=L+1} \mathbb{E}_{T \sim P(\textcolor{red}{\textbf{T}})} [Log (P(x_u | \textcolor{red}{\textbf{T}}))] - KL(p(\textcolor{red}{\textbf{T}}) || q(\beta))
    \label{equ:elbo}
\end{split}
\end{equation}

\textbf{Loss function of BPL} The new learning objective $P(X, \textcolor{red}{\textbf{T}}, \theta)$ over the whole data set has a supervised learning $P(X_L, \textcolor{red}{\textbf{T}}, \theta)$ which has not changed from Eq.~\ref{equ:pseudo_label_loss}, and an unsupervised learning part $P(X_U, \textcolor{red}{\textbf{T}}, \theta)$ from the above Eq.~\ref{equ:elbo}. The final loss function is an ELBO over the whole data set:

\begin{equation}
\begin{split}
    &\mathcal{L}^{VI}_{SegPL} = \underbrace{\frac{1}{L} \sum_{l=1}^{L} f_{dice}(\theta^{n-1}(x_l), y_l)}_{\mathcal{L}_L} + \\
    & \underbrace{\alpha \frac{1}{N-L} \sum_{u=L+1}^{N} f_{dice}(\theta^{n-1}(x_u), \mathds{1} (\theta^{n-1}(x_u) > \textcolor{red}{\textbf{T}}))}_{\mathcal{L}_U} + \\
    &\underbrace{Log(\sigma_\beta) - Log(\sigma) + \frac{\sigma^2 + (\mu-\mu_\beta)^2}{2*(\sigma_\beta)^2} - 0.5}_{\mathcal{L}_{KL}: \: KL(p(T) || q(\beta)), \: \beta \sim \mathcal{N}(\mu_\beta,\,\sigma_\beta)}
    \label{equ:variational_loss}
\end{split}
\end{equation}

\begin{figure*}[!h]
\centering
\includegraphics[width=0.9\textwidth]{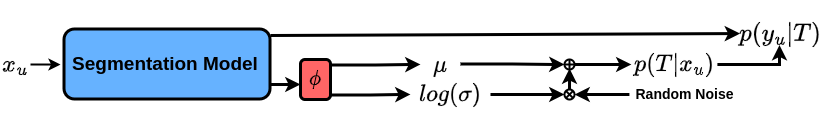}
\caption{The implementation of the proposed Bayesian Pseudo Labels. Only unsupervised learning part is illustrated.}
\label{fig:bpl_vae}
\end{figure*}

Where \textcolor{red}{\textbf{T}} can be found in Eq.~\ref{equ:reparam}. Different data sets might need different priors for the best empirical performances. Although we suggest to use higher mean such as 0.9 as a starting point. A schematic illustration of the implementation of Bayesian Pseudo Labels is shown in Fig. \ref{fig:bpl_vae}.


\section{Experimental Results}
\label{section:results}
\subsection{Data sets} 
\textbf{The classification of pulmonary arteries and veins (CARVE)} We use CARVE for demonstration of 3D binary segmentation of lung vessel of CT images. The CARVE dataset \cite{carve2014} comprises 10 fully annotated non-contrast low-dose thoracic CT scans. Each case has between 399 and 498 images, acquired at various spatial resolutions ranging from (282 x 426) to (302 x 474). We randomly select 1 case for labelled training, 2 cases for unlabelled training, 1 case for validation and the remaining 5 cases for testing. All image and label volumes were cropped to 176 $\times$ 176 $\times$ 3. To test the influence of the number of labelled training data, we prepared four sets of labelled training volumes with differing numbers of labelled volumes at: 2, 5, 10, 20. Normalisation was performed at case wise. Data curation resulted in 479 volumes for testing, which is equivalent to 1437 images. No data augmentation is used.

\textbf{BRATS 2018}  We use BRATS 2018 \cite{brats2015} \cite{DBLP:journals/corr/abs-1811-02629} for demonstration of 2D multi-class segmentation of brain tumour of MRI images. The BRATS 2018 comprises 210 high-grade glioma and 76 low-grade glioma MRI cases. Each case contains 155 slices. We focus on multi-class segmentation of sub-regions of tumours in high grade gliomas (HGG). All slices were centre-cropped to 176 x 176. We prepared three different sets of 2D slices for labelled training data: 50 slices from one case, 150 slices from one case and 300 slices from two cases. We use another 2 cases for unlabelled training data and 1 case for validation. 50 HGG cases were randomly sampled for testing. Case-wise normalisation was performed and all modalities were concatenated. A total of 3433 images were included for testing. No data augmentation is used.

\textbf{Task01 Brain Tumour} We use Task01 Brain Tumour from Medical Segmentation Decathlon consortium \cite{med_decathlon} as a demonstration of 3D binary segmentation of brain tumour of MRI images. The Task01 Brain Tumour is based on BRATS 2017 with different naming format from BRATS 2018. This data set was not in our previous MICCAI version but we included this data set here because it is easy to download and use for the readers for the future follow-up works. Each case in The Task01 Brain Tumour has 155 slices with 240 x 240 spatial dimension. We merge all of the tumour classes into one tumour class for simplicity. We do not apply centre cropping in the pre-processing here. In the training, we randomly crop volumes on the fly with size of 64 x 64 x 64. We separate the original training cases as labelled training data and testing data. We use the original testing cases as unlabelled data. For the labelled training data, we use 8 cases with index number from 1 to 8. We have 476 cases for testing and 266 cases for unlabelled training data. We apply normalisation with statistics of intensities across the whole training data set. We keep all of the MRI modalities as 4 channel input.

\textbf{Task05 Prostate} We also use Task05 Prostate from Medical Segmentation Decathlon consortium \cite{med_decathlon} as another demonstration of 3D binary segmentation of prostate of MRI images. Each case in the Task05 Prostate is a 4D volume: 2 modalities, 15 slices with 320 $\times$ 320 spatial dimension. We divide the original training cases into three parts, 1 case as labelled training data, 16 cases as unlabelled training data and the rest 14 cases as unseen testing data. During training, we randomly crop volumes on the fly with a target size of 192 $\times$ 192 $\times$ 8. 

\subsection{Baselines}
Our baselines include both supervised and semi-supervised learning methods. We use U-net~\cite{Unet} in SegPL as an example of segmentation network. Partly due to computational constraints, for 3D experiments we used a 3D U-net with 8 channels in the first encoder such that unlabelled data can be included in the same batch. For 2D experiments, we used a 2D U-net with 16 channels in the first encoder. The first baseline utilises supervised training on the backbone and is trained with labelled data denoted as ``Sup''. We compared SegPL with state-of-the-art consistency based methods: 1) ``cross pseudo supervision'' or CPS \cite{cpl_cvpr_2021}, which is considered the current state-of-the-art for semi-supervised segmentation; 2) another recent state-of-the-art model ``cross consistency training'' \cite{cct_cvpr2020}, denoted as ``CCT'', due to hardware restriction, our implementation shares most of the decoders apart from the last convolutional block; 3) a classic model called ``FixMatch'' (FM) \cite{fixmatch2020}. To adapt FixMatch for a segmentation task, we added Gaussian noise as weak augmentation and ``RandomAug'' \cite{random_augmentation} for strong augmentation; 4) ``self-loop \cite{self_loop_uncertainty}'', which solves a self-supervised jigsaw problem as pre-training and combines with pseudo-labelling.

\subsection{Training}
We use Adam optimiser\cite{adam} with default settings. Our code is implemented using Pytorch 1.0\cite{pytorch} and released in \url{https://github.com/moucheng2017/EMSSL}. We trained all of the experiments with a TITAN V GPU with 12GB memory. The training hyperparameters are included in Table \ref{tab:hyperparameters}. The prior values used are presented in Table \ref{tab:exps_prior_exps}.
\begin{table}[!h]
\caption{Hyper-parameters used across experiments. Different data might need different $\alpha$. LR: learning rate. Ratio: Unlabelled / labelled in each batch.}
    \centering
    \resizebox{0.48\textwidth}{!}
    {
    \begin{tabular}{c c c c c c }
        \hline
        Data & Batch Size & L.R. & Steps & $\alpha$ & Ratio.\\
        \hline
        BRATS(2D)  & 2 & 0.03 & 200 & 0.05 & 5\\
        \hline
        CARVE(3D) & 2 & 0.01 & 800 & 1.0 & 4\\
        \hline
         Task01(3D) & 1 & 0.0004 & 25000 & 0.1 & 2\\
         \hline
         Task05(3D) & 1 & 0.001 & 2000 & 0.002 & 4 \\
         \hline
    \end{tabular}
    }
    \label{tab:hyperparameters}
\end{table}

\begin{table}[!h]
\caption{Different prior values of $T$ used across experiments.}
    \centering
    \resizebox{0.25\textwidth}{!}
    {
    \begin{tabular}{c c c }
        \hline
        Data & Mean & Std. \\
        \hline
        BRATS(2D)  & 0.5 & 0.1 \\
        \hline
        CARVE(3D) & 0.4 & 0.1 \\
        \hline
         Task01(3D) & 0.9 & 0.1 \\
         \hline
         Task05(3D) & 0.9 & 0.1 \\
         \hline
    \end{tabular}
    }
    \label{tab:exps_prior_exps}
\end{table}

\subsection{Segmentation performances}

\begin{figure}[!h]
    \centering
    \begin{center}
        \includegraphics[width=0.6\textwidth]{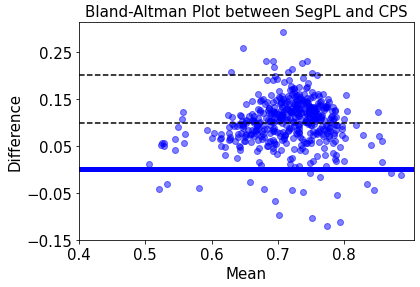}
    \end{center}
    \caption{SegPL statistically outperforms the best performing baseline CPS when trained on 2 labelled volumes from the CARVE dataset. Each data point represents a single testing image.}
    \label{fig:iou_analysis}
\end{figure}

\begin{table*}[!h]
\caption{Our model vs Baselines on a binary vessel segmentation task on 3D CT images of the CARVE dataset. Metric is Intersection over Union (IoU ($\uparrow$) in \%). Avg performance of 5 training. \textcolor{blue}{blue: 2nd best}. \textcolor{red}{red: best}}
  \label{tab:carve}
  \centering
  \resizebox{\textwidth}{!}
  {
  \begin{tabular}{ c  | c  | c | c | c | c | c}
    \hline
    \bfseries Data & \bfseries Supervised & \multicolumn{5}{c}{\bfseries Semi-Supervised} \\
    \hline
    \bfseries Labelled & \bfseries 3D U-net &\bfseries FixMatch  & \bfseries CCT &\bfseries CPS &\bfseries SegPL &\bfseries SegPL+VI\\
    \bfseries Volumes & \cite{Unet} & \cite{fixmatch2020} & \cite{cct_cvpr2020} & \cite{cpl_cvpr_2021} & (Ours, 2022) & (Ours, 2022)\\
    \hline
    2 & 56.79$\pm$6.44 & 62.35$\pm$7.87 & 51.71$\pm$7.31 & 66.67$\pm$8.16 & \textbf{\textcolor{blue}{69.44$\pm$6.38}} & \textbf{\textcolor{red}{70.65$\pm$6.33}}\\
    \hline
    5 & 58.28$\pm$8.85 & 60.80$\pm$5.74 & 55.32$\pm$9.05 & 70.61$\pm$7.09 & \textbf{\textcolor{red}{76.52$\pm$9.20}} & \textbf{\textcolor{blue}{73.33$\pm$8.61}}\\
    \hline
    10 & 67.93$\pm$6.19 & 72.10$\pm$8.45 & 66.94$\pm$12.22 & 75.19$\pm$7.72 & \textbf{\textcolor{blue}{79.51$\pm$8.14}} &\textbf{\textcolor{red}{79.73$\pm$7.24}}\\
    \hline
    20 & 81.40$\pm$7.45 & 80.68$\pm$7.36 & 80.58$\pm$7.31 & 81.65$\pm$7.51 & \textbf{\textcolor{blue}{83.08$\pm$7.57}} &\textbf{\textcolor{red}{83.41$\pm$7.14}}\\
    \hline
    \multicolumn{7}{c}{\bfseries Computational need} \\
    \hline
    Train(s) & 1014 & 2674 & 4129 & 2730 & 1601 & 1715\\
    \hline 
    Flops & 6.22 & 12.44 & 8.3 & 12.44 & 6.22 & 6.23\\
    \hline
    Para(K) & 626.74 & 626.74 & 646.74 & 1253.48 & 626.74 & 630.0 \\
    \hline
    \end{tabular}
    }
\end{table*}

\begin{table*}[!h]
\caption{Our model vs Baselines on multi-class tumour segmentation on 2D MRI images of BRATS 2018. Metric is Intersection over Union (IoU ($\uparrow$) in \%). Avg performance of 5 runs. \textcolor{blue}{blue: 2nd best}. \textcolor{red}{red: best}}
  \label{tab:brats}
  \centering
  \resizebox{\textwidth}{!}
  {
  \begin{tabular}{ c  | c  |  c | c | c | c | c}
    \hline
    \bfseries Data & \bfseries Supervised & \multicolumn{5}{c}{\bfseries Semi-Supervised} \\
    \hline
    \bfseries Labelled & \bfseries 2D U-net &\bfseries Self-Loop &\bfseries FixMatch &\bfseries CPS &\bfseries SegPL &\bfseries SegPL+VI\\
    \bfseries Slices & \cite{Unet} & \cite{self_loop_uncertainty} & \cite{fixmatch2020} & \cite{cpl_cvpr_2021} & (Ours, 2022) & (Ours, 2022)\\
    \hline
    50 & 54.08$\pm$10.65 & 65.91$\pm$10.17 & 67.35$\pm$9.68 & 63.89$\pm$11.54 & \textcolor{blue}{\textbf{70.60$\pm$12.57}} &\textcolor{red}{\textbf{71.20$\pm$12.77}}\\
    \hline
    150 & 64.24$\pm$8.31 & 68.45$\pm$11.82 & 69.54$\pm$12.89 & 69.69$\pm$6.22 & \textcolor{blue}{\textbf{71.35$\pm$9.38}} & \textcolor{red}{\textbf{72.93$\pm$12.97}}\\
    \hline
    300 & 67.49$\pm$11.40 & 70.80$\pm$11.97 & 70.84$\pm$9.37 & 71.24$\pm$10.80 & \textcolor{blue}{\textbf{72.60$\pm$10.78}} & \textcolor{red}{\textbf{75.12$\pm$13.31}}\\
    \hline
    \end{tabular}
    }
\end{table*}

\begin{table}[!h]
\caption{Our model vs Supervised baseline on 3D binary tumour segmentation of Task 01 Brain Tumour (BRATS 2017). Metric is Intersection over Union (IoU ($\uparrow$) in \%). Avg performance of models between iteration 20000 and 25000 with 1000 as the interval. The shape of each training sample is $96^3$. \textcolor{red}{red: best}}
    \centering
    \resizebox{0.6\textwidth}{!}
    {
    \begin{tabular}{c c c c }
        \hline
        Testing size & 32 $\times$ 32 $\times$ 32 & 64 $\times$ 64 $\times$ 64 & 96 $\times$ 96 $\times$ 96 \\
        \hline
        Supervised  & 61.07$\pm$7.93 & 66.94$\pm$12.4 & 70.13$\pm$13.22  \\
        \hline
        SegPL-VI & \textcolor{red}{64.44$\pm$8.3} & \textcolor{red}{71.43$\pm$11.91} & \textcolor{red}{73.07$\pm$11.71} \\
        \hline
    \end{tabular}
    }
    \label{tab:results_brats2017}
\end{table}

\begin{table}[!h]
\caption{Our model vs Supervised baseline on 3D binary segmentation of prostate Task 05 from Medical Decathlon. Metric is Intersection over Union (IoU ($\uparrow$) in \%). Performance of the models which achieved the highest training accuracy. The shape of each training sample is 192 $\times$ 192 $\times$ 8. \textcolor{red}{red: best}}
    \centering
    \resizebox{0.49\textwidth}{!}
    {
    \begin{tabular}{c c c c }
        \hline
        Testing size & 192 $\times$ 192 $\times$ 8 & 160 $\times$ 160 $\times$ 8 & 128 $\times$ 128 $\times$ 8 \\
        \hline
        Supervised  & 67.68$\pm$10.06 & 61.39$\pm$11.86 & 60.53$\pm$9.94 \\
        \hline
        SegPL-VI & \textcolor{red}{70.15$\pm$10.59} & \textcolor{red}{63.15$\pm$11.62} & \textcolor{red}{61.06$\pm$10.39} \\
        \hline
    \end{tabular}
    }
    \label{tab:results_task05}
\end{table}

\begin{figure*}[!h]
   \begin{minipage}{0.445\textwidth}
     \centering
     \includegraphics[width=\linewidth]{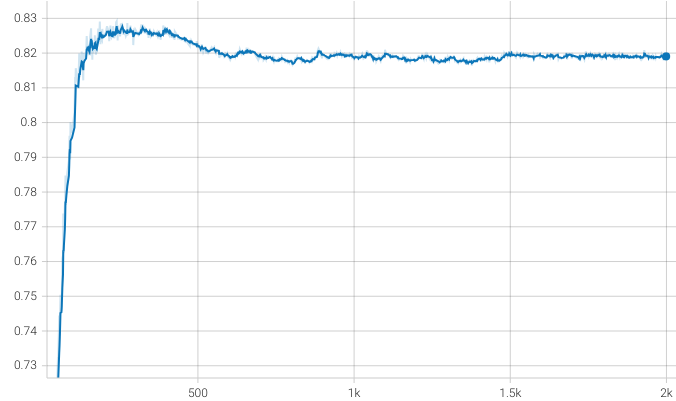}
         \caption{Y-axis: Learnt threshold in the experiment of Task01 Brain Tumour. X-axis: training iterations. The mean of the prior is 0.9 and the std of the prior is 0.1. The learnt threshold converged around 0.82 after 2000 iterations.}\label{fig:task01_threshold}
   \end{minipage}\hfill
   \begin{minipage}{0.49\textwidth}
     \centering
     \includegraphics[width=\linewidth]{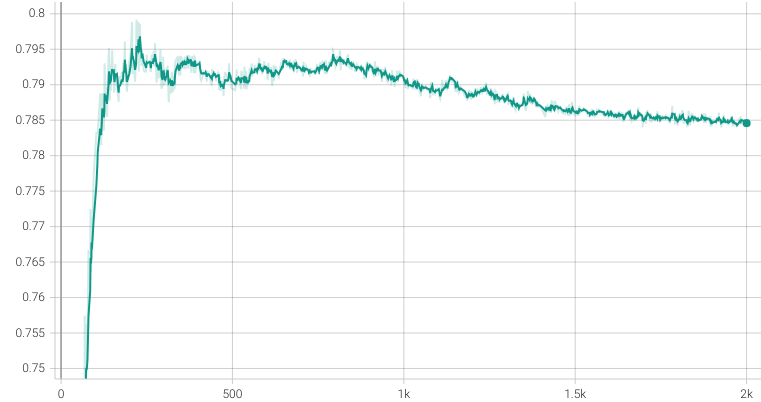}
     \caption{Y-axis: Learnt threshold in the experiment of Task05 Prostate. X-axis: training iterations. The mean of the prior is 0.9 and the std of the prior is 0.1. The learnt threshold converged around 0.785 after 2000 iterations.}\label{fig:task05_threshold}
   \end{minipage}
\end{figure*}

\begin{figure}[!h]
\centering
\includegraphics[width=0.9\textwidth]{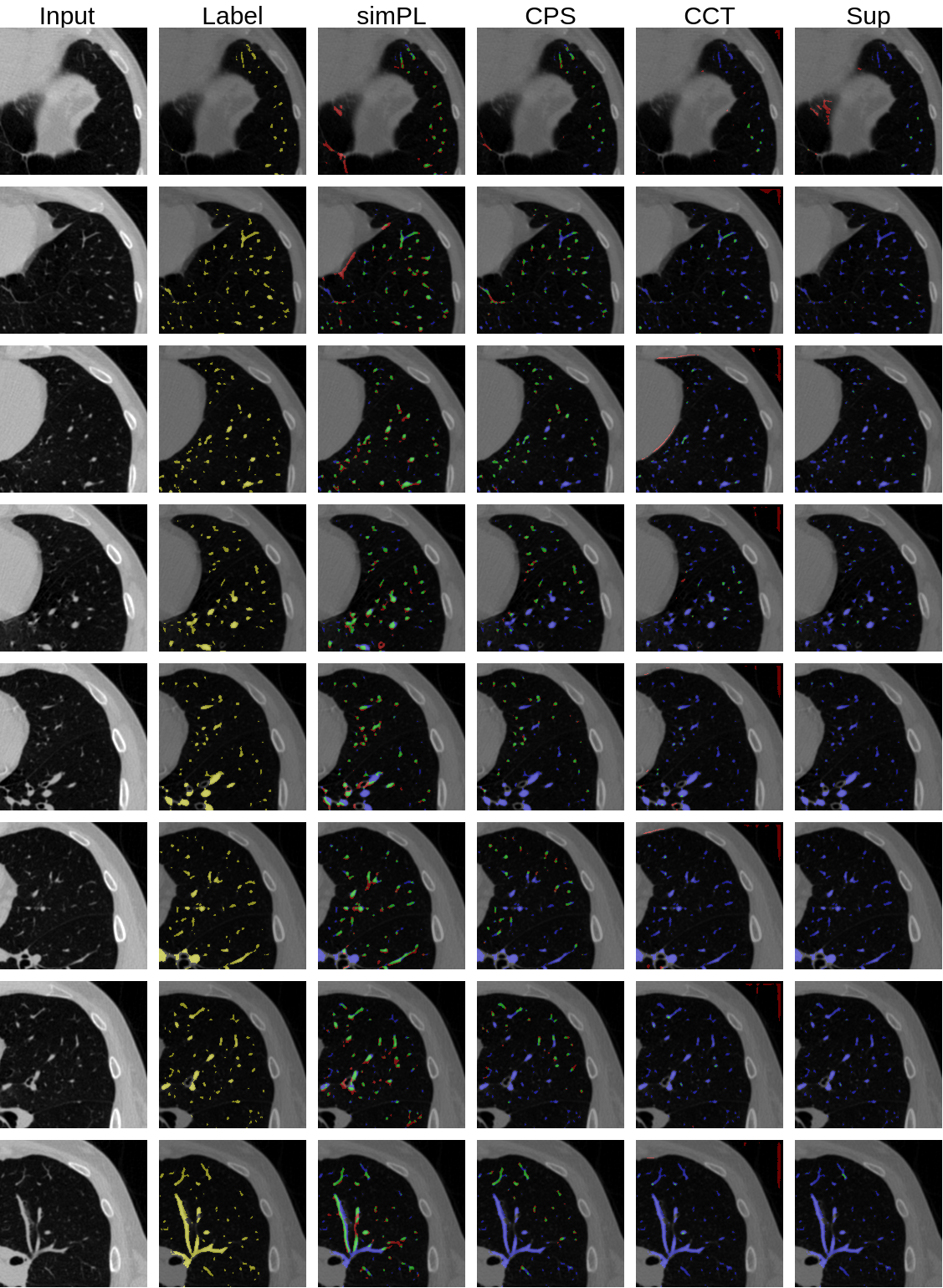}
\caption{Visual results. CARVE trained with 5 labelled volumes. Red: false positive. Green: true positive. Blue: false negative. Yellow: ground truth. GT: Ground truth. CPS: cross pseudo labels (CVPR 2021). CCT: cross consistency training (CVPR 2020). Sup: supervised training.}
\label{fig:visual_results_carve}
\end{figure}

\begin{figure}[!h]
\centering
\includegraphics[width=0.9\textwidth]{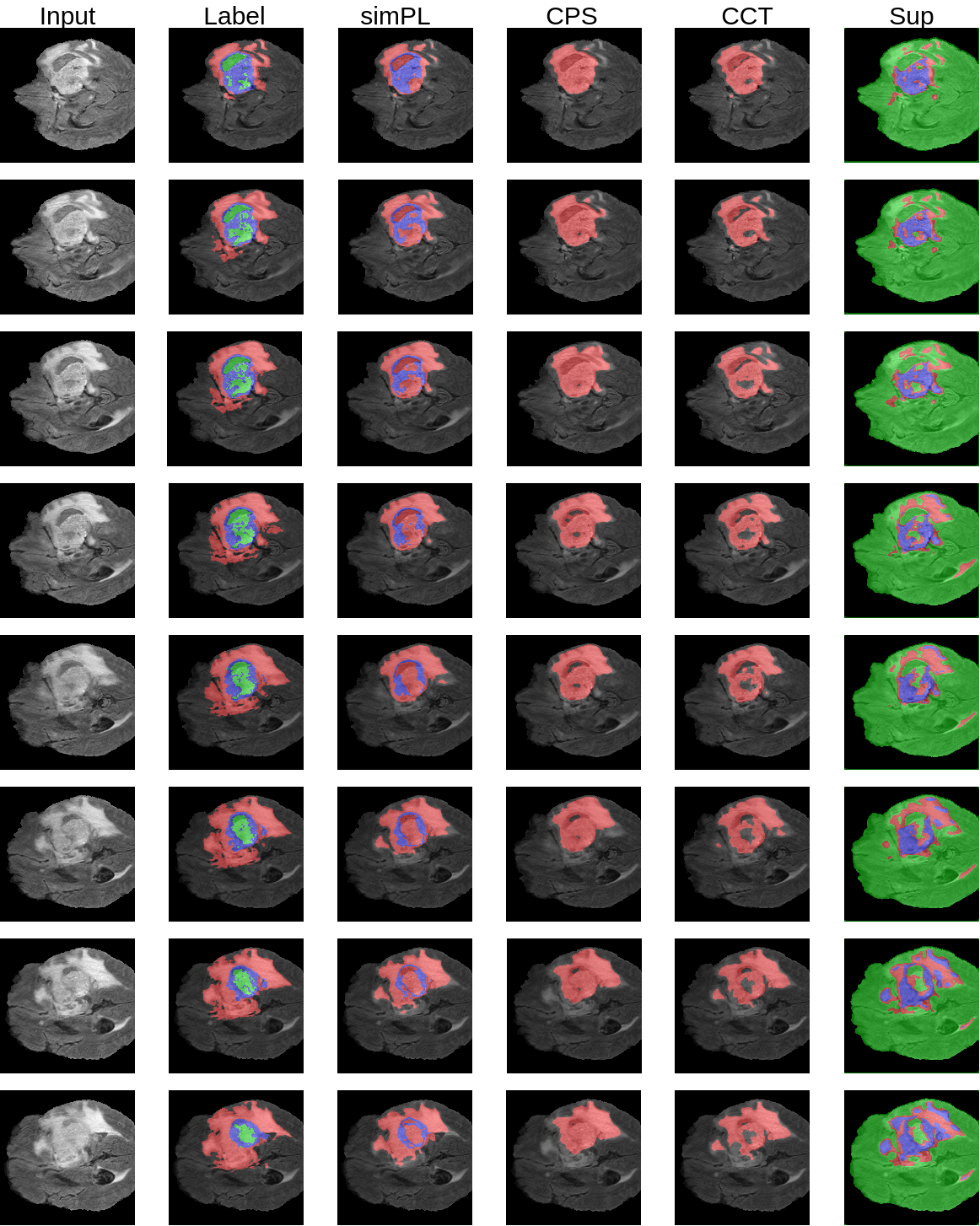}
\caption{Visual results. BRATS 2018 trained with 300 labelled slices. Red: whole tumour. Green: tumour core. Blue: enhancing tumour core. GT: Ground truth. CPS: cross pseudo labels (CVPR 2021). CCT: cross consistency training (CVPR 2020). Sup: supervised training.}
\label{fig:visual_results_brats}
\end{figure}

The segmentation performances of CARVE 2014, BRATS 2018, Task 01 can be found in Tab.\ref{tab:carve},  Tab.\ref{tab:brats} and Tab.\ref{tab:results_brats2017}, respectively. As reflected in the quantitative results in tables, pseudo labelling based SegPL consistently achieves better results than the baselines of semi-supervised and supervised methods. Especially, as shown in Fig.\ref{fig:iou_analysis} of the Bland-Altman plot between the best performing baseline CPS and our SegPL on CARVE when only 2 labelled volumes are used for training, SegPL statistically outperforms the best baseline. We further confirm the statistical difference by performing Mann Whitney test on the same results on 2 labelled volumes and we found the p-vale less than 1e-4. By extending the SegPL with variational inference to SegPL-VI, we found further improvements on segmentation on most of the experiments. Interestingly, the improvements brought by SegPL-VI is more obvious on multi-class experiments on BRATS 2018. As the outputs on BRATS are multi-channel but SegPL-VI learns one threshold across all of the channel, we suspect that might bring in strong regularisation effect which results in noticeable improvements. We also noticed that SegPL-VI could fail to learn optimal threshold sometimes as the result of SegPL-VI on CARVE with 5 labelled volumes are inferior to the corresponding result of SegPL. We expect that more hyper-parameter searching could improve the performance of SegPL.

As shown in the qualitative results in Fig\ref{fig:visual_results_carve} of CARVE, SegPL successfully learnt better decision boundary than other baselines that SegPL can partially separate the foreground lung vessels from the background whereas most of the other methods classifies everything as background. However, SegPL seemed to have overconfident predictions on the edges of the foreground that it has a lot of false positive results. Similarly in BRATS, SegPL detected one more class of brain tumour (blue) than the other baselines in Fig\ref{fig:visual_results_brats}. However, none of the methods including SegPL can detect the most rare green class of tumour. 

One phenomenon worthy mentioning is shown in Tab.\ref{tab:results_brats2017} on 3D binary segmentation of whole tumour and Tab.\ref{tab:results_task05} on 3D binary segmentation of prostate. During training on whole brain tumour segmentation, we use random cropping with fixed size at 64 x 64 x 64 to compensate with the memory of GPU. On testing data, we examined the models with different sizes of cropped volumes at $32^3$, $64^3$, $96^3$ and $128^3$. The models actually generalise well on the scales that they haven't seen during the training. In fact, larger cropped volumes result in better results. The results in Tab.\ref{tab:results_task05} on segmentation of prostate also confirms this phenomenon.

Although SegPL achieves higher segmentation accuracy, SegPL enjoys a low computational burden. As illustrated in the computational need section in Tab.\ref{tab:carve}, SegPL has the least computational burden among all of the tested semi-supervised learning baselines. Especially in terms of FLOPs, SegPL is very close to supervised learning methods. This shows that our model has the scaling potential for large models and large data sets.

\subsection{Sensitivity Studies of Hyper-parameters}
\begin{figure*}[!ht]
\centering
\includegraphics[width=\textwidth]{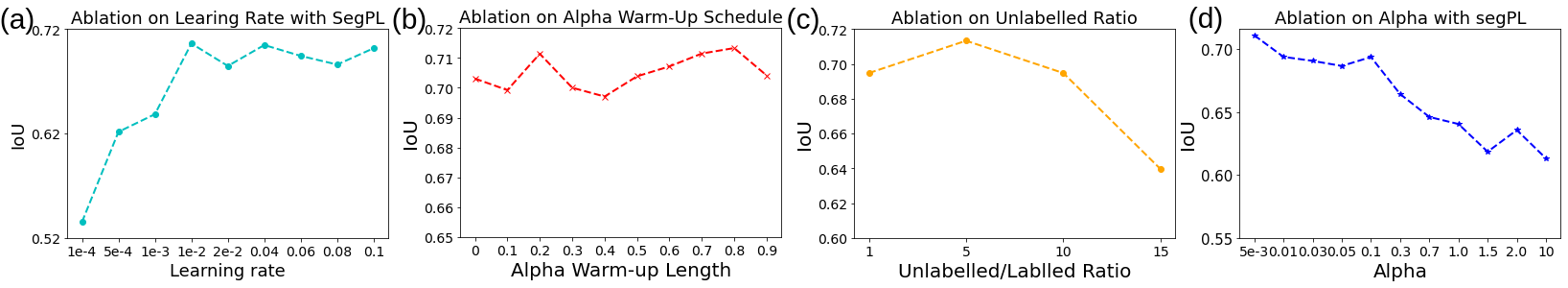}
\caption{Sensitivity Studies of Hyper-parameters on BRATS with 150 labelled slices.}
\label{fig:ablation}
\end{figure*}

We performed brief sensitivity studies on hyper-parameters on BRATS with 150 labelled slices. As shown in Fig.\ref{fig:ablation}, a) shows that SegPL is very sensitive to learning rate that it should be at least 0.01. We found that other baselines also needed large learning rate. Fig.\ref{fig:ablation}.b) shows the impact of warm-up schedule of $\alpha$ from 0 to final $\alpha$ value. x axis is the length of linear warming-up of $\alpha$ in terms of whole steps. It appears that SegPL is not sensitive to warm-up schedule of $\alpha$. Fig.\ref{fig:ablation}.c) illustrates the effect of the ratio between unlabelled images to labelled images in each batch. The suitable range of unlabelled/labelled ratio is quite wide and between 1 to 10. Fig.\ref{fig:ablation}.d) shows that the pseudo supervision cannot be too strong. This confirms the suggestions from the original pseudo labelling paper that pseudo supervision should not dominate the training.


\subsection{Robustness}
\begin{figure*}[!h]
   \begin{minipage}{0.49\textwidth}
     \centering
     \includegraphics[width=\linewidth]{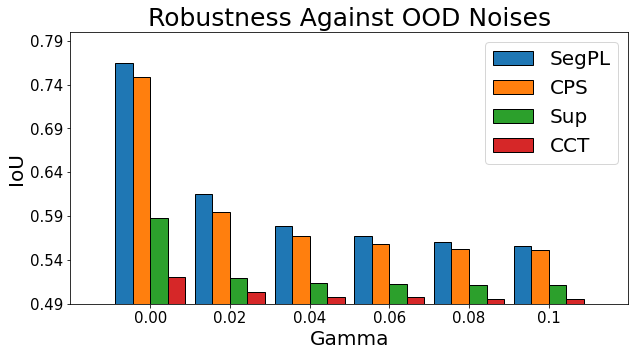}
         \caption{Robustness against out-of-distribution noise. Gamma is the strength of the out-of-distribution noises. Using 2 labelled volumes from CARVE.}\label{fig:out_of_distribution}
   \end{minipage}\hfill
   \begin{minipage}{0.49\textwidth}
     \centering
     \includegraphics[width=\linewidth]{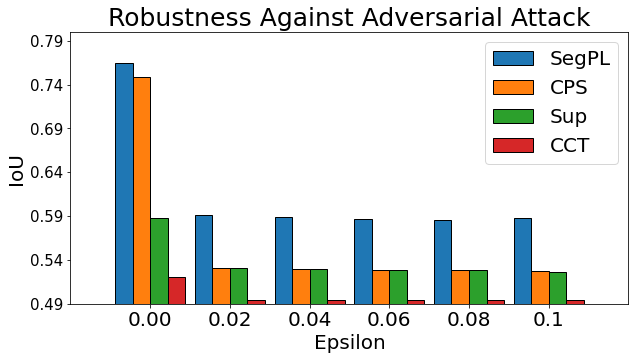}
     \caption{Robustness against adversarial attack. Epsilon is the strength of the FGSM\cite{adversarial_attack} attack. Using 2 labelled volumes from CARVE.}\label{fig:adversarial_attack}
   \end{minipage}
\end{figure*}

In medical imaging, models often face challenges due to out-of-distribution (OOD) noise such as variations in scan acquisition parameters or differing patient populations. These factors can significantly degrade model performance in real-world applications. To evaluate the robustness of our proposed SegPL model against OOD noise, we conduct experiments using models trained on the CARVE dataset. 

We simulate OOD noises with unseen random contrast and Gaussian noise, we then apply mix-up \cite{mixup} to create new testing samples by adding the OOD noises on original images. Specifically, for a given original testing image $x_t$, we applied random contrast and noise augmentation on $x_t$ to derive OOD samples $x'_t$. We arrived at the testing sample ($\hat{x_t}$) via $\gamma x'_t + (1 - \gamma) x_t$. As shown in Fig.\ref{fig:out_of_distribution}, as testing difficulty increases, the performances across all baselines drop exponentially. SegPL outperformed all of the baselines across all of the tested experimental settings. The findings suggest that SegPL is more robust when testing on OOD samples and achieves better generalisation performance against that from the baselines. 

In the context of privacy and security, especially as federated learning across hospitals gains popularity, the robustness against adversarial attacks becomes crucial. We assess SegPL's resilience to such attacks using the fast gradient sign method (FGSM) \cite{adversarial_attack}. FGSM perturbs an image by computing the gradient of the loss function with respect to the input image and adding a noise term proportional to the sign of the gradient.

Our experiments show that the performance of all models, including SegPL, declines as the strength of the adversarial attack (measured by Epsilon) increases. However, SegPL exhibits a smaller drop in performance compared to baseline models, as illustrated in Fig.\ref{fig:adversarial_attack}. These results further substantiate the robustness of SegPL under various challenging conditions.


\subsection{Uncertainty}
Since SegPL-VI is trained with stochastic threshold for unlabelled data therefore not suffering from posterior collapse. Consequently, SegPL can generate plausible segmentation during inference using stochastic thresholds. To test the performance of SegPL-VI on uncertainty quantification, we use random latent variable values (threshold) with 5 Monte Carlo samples. We focus experimenting on models trained with 5 labelled volumes of CARVE data set. For comparison, we adopt Deep Ensemble, as it is the gold-standard baseline for uncertainty estimation \cite{deep_ensemble}\cite{evaluation_uncertainty}. Both the tested methods Deep Ensemble and SegPL-VI achieved the same Brier score at 0.97. This result shows that SegPL-VI has the potential to become a benchmark method for uncertainty quantification. The Brier score is calculated using beneath equation, where, $y_{ij}$ is the ground truth label at pixel at location i, j, $y_{ij}$ is 1 for foreground pixel and $y_{ij}$ is 0 for background pixel. $p_{ij}$ is the predicted probability of the pixel being the foreground pixel.
\begin{equation}
\label{brier}
Brier = \frac{1}{HW} \sum_{i=1}^{H} \sum_{j=1}^{W} (p_{ij} - y_{ij})^2
\end{equation} 

\section{Limitations and future works}
There are two main limitations of the proposed Bayesian pseudo labels. The first limitation is that once the model starts to over-fit, the model becomes overconfident that it predicts with very high confidence, while the learnt threshold also converges to a value such as shown in Fig.\ref{fig:task01_threshold} and Fig.\ref{fig:task05_threshold}. In this situation, if the prior of the mean is too low, then the learnt threshold won't be able to mask out the bad over confident pseudo labels. Thus calibration becomes very important here. In future work, one could extend the formulation of pseudo labels to take into account of calibration. 

The second limitation is the use of the prior in the current paper. We use Gaussian due to its simplicity and easy to implement. However, Gaussian prior might not be the most optimal one here. Future work can explore the impact of other priors of learnt threshold. Candidate prior distributions include categorical and Beta distributions. 

In terms of implementation, the current Bayesian Pseudo Labels only learns a single threshold for all of the images in the same batch size. However, more adaptive implementations could potentially boost the performance of Bayesian Pseudo Labels, such as learning one threshold for each image or even each pixel. If the thresholds are learnt per pixel of an image, one might also need to consider the spatial correlations among the thresholds.

Theoretically, another interesting future work can be studying the impact of labelled data in terms of preventing collapsed representations. Other future work can also look into the convergence property of SegPL-VI.

The feasibility of the applications of the proposed methods on other tasks such as uncertainty quantification, classification and registration also remain unexplored. 

In the future pipeline for learning with limited annotations, we also expect to exploit SegPL-VI's full potential by combining with large-scale pre-training techniques.




\section{Conclusions}
In this paper, we revisit pseudo-labelling and provide an interpretation of its empirical success by formulating the pseudo-labelling process as the EM algorithm. We as well unravel its full formulation along with a learning based approach to approximate it. Empirically, we examined that the original pseudo-labelling \cite{PseudoLabel} and its Bayesian generalisation on semi-supervised medical image segmentation and we report that pseudo-labelling as a competitive and robust baseline. 

\section{Acknowledgements}
MCX was supported by GSK (BIDS3000034123) and UCL Engineering Dean’s Prize. NPO is supported by a UKRI Future Leaders Fellowship (MR/S03546X/1). DCA is supported by UK EPSRC grants M020533, R006032, R014019, V034537, Wellcome Trust UNS113739, Wellcome Trust 221915/Z/20/Z. JJ is supported by Wellcome Trust Clinical Research Career Development Fellowship 209,553/Z/17/Z. NPO, DCA, and JJ are supported by the NIHR UCLH Biomedical Research Centre, UK. This research was funded in whole or in part by the Wellcome Trust [209553/Z/17/Z]. For the purpose of open access, the author has applied a CC-BY public copyright licence to any author accepted manuscript version arising from this submission.
\printbibliography
\end{document}